\documentclass{Interspeech2024}
\usepackage{multirow}
\usepackage{makecell}
\usepackage{fdsymbol}
\usepackage{balance}

\interspeechcameraready

\title{Diffusion Synthesizer for Efficient Multilingual Speech to Speech Translation}

\name{Nameer}{Hirschkind}
\name{Xiao}{Yu}
\name{Mahesh Kumar}{Nandwana}
\name{Joseph}{Liu}
\name{Eloi}{DuBois}
\name{Dao}{Le}
\name{Nicolas}{Thiebaut}
\name{Colin}{Sinclair}
\name{Kyle}{Spence}
\name{Charles}{Shang}
\name{Zoe}{Abrams}
\name{Morgan}{McGuire}

\address{
  Roblox, United States}
\email{nhirschkind@roblox.com}  


\begin{document}
\maketitle 

\begin{abstract}
    
We introduce DiffuseST, a low-latency, direct speech-to-speech translation system capable of preserving the input speaker's voice zero-shot while translating from multiple source languages into English. We experiment with the synthesizer component of the architecture, comparing a Tacotron-based synthesizer to a novel diffusion-based synthesizer. We find the diffusion-based synthesizer to improve MOS and PESQ audio quality metrics by 23\% each and speaker similarity by 5\% while maintaining comparable BLEU scores. Despite having more than double the parameter count, the diffusion synthesizer has lower latency, allowing the entire model to run more than 5$\times$ faster than real-time.
\end{abstract}

\section{Introduction}
Speech-to-speech translation (S2ST) has the potential to transform the way we communicate with others who do not speak the same language. The simplest way to perform S2ST automatically is to chain automatic speech recognition (ASR) with text-to-text machine translation (MT) followed by text-to-speech synthesis (TTS), in what is called a ``cascaded" system \cite{seamless}. However, cascaded systems are slow and do not take full advantage of non-textual information imparted by the audio modality. For example, the tone of the input audio could help choose between several words with meanings close to "sorry" in the target language. In recent years, models that translate directly to speech in the target language and can be optimized end-to-end have outperformed cascaded systems \cite{seamless, translatotron2, unity, audiopalm}. These works rely on intermediary representations of text in the target language, such as discrete acoustic units or phonemes \cite{translatotron2, unity}. Systems like AudioPalm and VioLA use a single decoder-only transformer architecture \cite{audiopalm, viola}, while others like SeamlessM4T, UniTY, and Translatotron2 use separate encoder, decoder, and synthesizer components \cite{seamless, translatotron2, unity}. In this work, we study direct S2ST systems with separate encoder, decoder, and synthesizer modules.

To make translated communication more natural, much work has been done to enable S2ST systems to output speech in the same voice, emotion, and prosody as their input \cite{translatotron2, viola, seamlessex, translatotron3, styles2st}. This is known as zero-shot voice cloning or speaker preservation. Some prior works add architectural components separately trained to capture speaker characteristics \cite{seamlessex, styles2st}. Others attempt to capture these characteristics implicitly while training on pairs of utterances with the same speaking style and expression \cite{translatotron2, viola, translatotron3}. We take this approach but improve on prior works by pretraining the synthesizer on diverse voices to enable our model to learn speaker preservation with less data.

To make S2ST systems as close to real monolingual interaction as possible, it is critical to run at low latency in a streaming context (i.e. the system starts speaking before the input speaker is done talking). The rare S2ST systems addressing this challenge still operate with over 2.5 seconds of ending delay \cite{seamlessex}. While we do not tackle streaming directly in this work, we believe an important step to reduce system delay is to reduce model latency via parameter-efficient, low-latency, direct S2ST.

Developments in direct S2ST research are heavily influenced by work on TTS systems \cite{seamless, translatotron2, viola}.  Recently, diffusion models such as Voicebox and NaturalSpeech2 have been shown to work well for speech synthesis \cite{voicebox, naturalspeech2}. Diffusion models are attractive because they can produce diverse audio, pretrain on unlabeled audio data, and run non-autoregressively \cite{voicebox, naturalspeech2, flowmatching, audioldm}. Despite their great potential, existing works have yet to use such audio diffusion models for S2ST.

We introduce DiffuseST, a direct S2ST system that translates from many languages into English with a novel diffusion-based synthesizer that can perform low-latency S2ST with speaker preservation given as few as 3 seconds of input audio. The main contributions of this work are as follows: 1) We are the first work we know of to use a diffusion synthesizer for S2ST. We show our synthesizer is capable of zero-shot speaker preservation using implicit extraction of speaker characteristics, improving audio quality metrics by 23\% and speaker similarity by 5\% over a baseline. 2) We show the feasibility of S2ST with a much smaller architecture than previous works, enabling $5\times$ faster than real-time inference and facilitating future work on streaming. 3) We are one of the first S2ST works to rely only on public data while still training on over 1k hours of audio, making our work large-scale but more reproducible than prior research.

\section{Related Work}
In this section, we review prior S2ST and TTS research most relevant to our work. We draw the most architectural inspiration from Translatotron2 and Seamless, both direct, zero-shot voice cloning S2ST models \cite{seamless, translatotron2, seamlessex}. Both train on artificially-generated speaker-preserving label audios generated by TTS systems, but Seamless also trains on automatically aligned utterances mined from a massive multilingual corpus using the SONAR algorithm \cite{sonar}. While Seamless uses a separate expressivity module to capture speaker characteristics, Translatotron2 does so implicitly using an attention mechanism.

Other S2ST systems such as VioLA or AudioPaLM use LLM-style architectures to perform S2ST by predicting both speech and text tokens with a single autoregressive transformer \cite{audiopalm, viola}. These models demonstrate good quality but tend to have high parameter counts and can only run auto-regressively, hurting latency.

Several recent works in TTS have used diffusion with great success. Voicebox is such a model that leverages a flow-matching objective to learn to generate Mel spectrograms given context audio and text \cite{voicebox, flowmatch}. Recent work in using auto-encoders to learn discrete audio representations (e.g. EnCodec and SoundStream) enables performing audio diffusion in latent space \cite{encodec, soundstream}. NaturalSpeech2 is such a diffusion-style TTS model that uses continuous embeddings of discrete acoustic tokens as targets, rather than Mel spectrograms \cite{naturalspeech2}. Waveforms are then generated based on these latent features using the decoder of the same auto-encoder as a vocoder (module that produces waveforms given representations). SpeechFlow is a foundational model for speech generation tasks that uses Voicebox's flow-matching objective to generate EnCodec tokens \cite{flowmatching}. After pretraining on a large corpus of unlabeled audio data, SpeechFlow can be fine-tuned to perform downstream tasks like audio quality enhancement or TTS.

\section{System Description}
\begin{figure}[t]
    \centering
    \includegraphics[width=\linewidth]{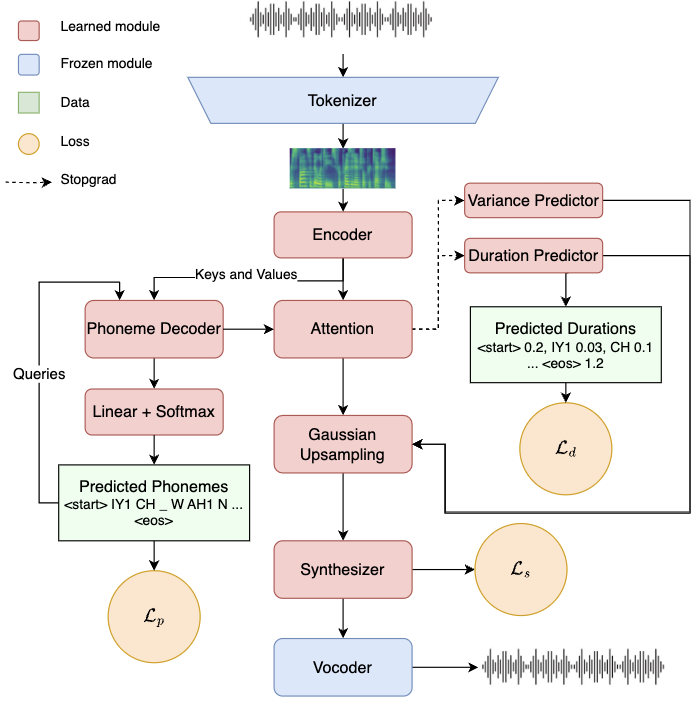}
    \caption{{Architecture Diagram of the proposed DiffuseST Model.}}
    \label{fig:whole}

    \vspace{-15pt}
\end{figure}

In this section, we characterize our novel S2ST system, DiffuseST, delving into model architecture, training, and inference strategies. We give particular attention to the diffusion synthesizer as well as other parts of our architecture that differ significantly from prior work.

\subsection{Model Architecture}
  As shown in Figure \ref{fig:whole}, DiffuseST comprises a tokenizer, acoustic encoder, phoneme decoder, upsampling and duration prediction module, synthesizer, and vocoder \cite{seamless, translatotron2}. To facilitate real-time use-cases later on, we target small parameter counts for all components. 
 
 The tokenizer is a waveform to Mel Spectrogram converter that runs on GPU. For the acoustic encoder, we use the Whisper-Small encoder due to its high quality pretraining \cite{whisper}. For the phoneme decoder, we use a transformer decoder with rotary embeddings and a softmax head to predict phonemes in the target language given context from the encoder \cite{aiayn, rope}. We apply a cross-entropy loss on the phoneme predictions $\mathcal{L}_{p}$. Inspired by Translatotron2, we add a single multi-headed cross-attention layer—with no masking or autoregression—that uses the decoder hidden states as queries and the encoder outputs as keys and values \cite{translatotron2}. This gives the model a second chance to extract features from the input audio that may be useful for voice cloning in the synthesizer.

 We use two small transformer encoders with output dimension 1 and softplus activation to predict the durations and variances of each phoneme given the outputs of the attention layer. We supervise the duration predictor directly with ground truth, per-phoneme durations via an $L_2$ loss $\mathcal{L}_{d}$. Variances are supervised implicitly by the final loss term from the synthesizer. We stop gradients before both the duration and variance predictors to improve training stability. We then use the Gaussian upsampling method from Non-Attentive Tacotron (NAT) to upsample the attention layer outputs to the frequency required by the synthesizer \cite{nat}. At train time we use ground truth durations for upsampling, but we always use the predicted variances, allowing the variance predictor to receive gradients from the synthesizer.

 The upsampled representations and durations contain all the information necessary to speak the predicted phonemes in the same voice as the input. This makes our architecture highly adaptable in that different synthesizers can be swapped out on top of the upsampling layer. We experimented with several synthesizer options and found a diffusion-based synthesizer to yield the best audio quality and speaker preservation. However, as the diffusion synthesizer is too unstable to learn unless the rest of the S2ST network weights are frozen, we first train all network parameters with a NAT-based synthesizer like Translatotron2 before replacing it with the diffusion synthesizer \cite{translatotron2, nat}, freezing all other parameters, and fine-tuning the synthesizer at the end of the training process. We refer to the loss term used to train the synthesizer generically as $\mathcal{L}_{s}$.

 \subsection{Diffusion Synthesizer}
\begin{figure}[b]
    \centering
    \includegraphics[width=\linewidth]{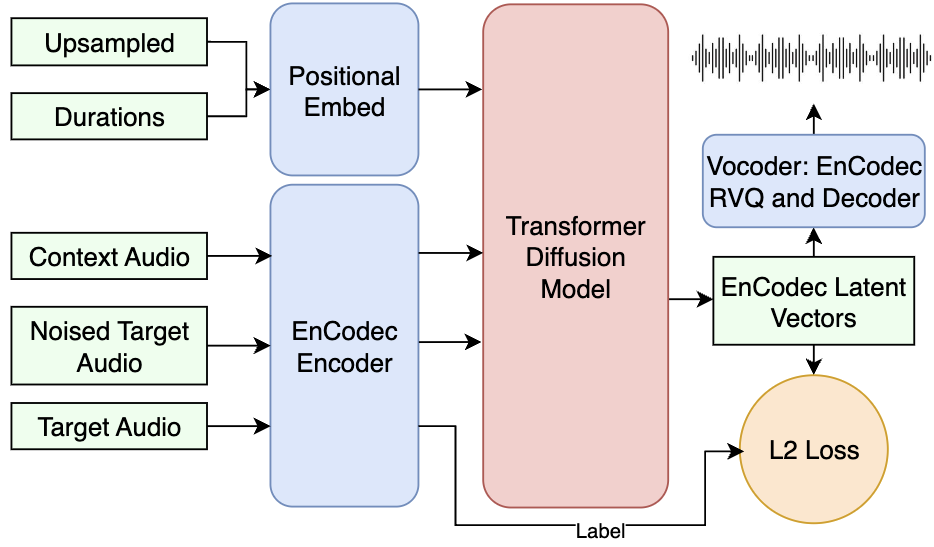}
    \caption{{Architecture of the Diffusion Synthesizer in the DiffuseST model.}}
    \label{fig:diff}

\end{figure}
 
 We will now go into greater depth on our novel diffusion synthesizer (see Figure \ref{fig:diff}). We base our work on an open source implementation of Voicebox \cite{voicebox}.\footnote{Credit to GitHub user lucidrains for the implementation} The aspect we are most interested in using from Voicebox is the conditional flow matching diffusion objective \cite{flowmatch}, as it is reported to require fewer diffusion steps at inference time compared to other diffusion objectives \cite{voicebox}. We do not use Voicebox's duration predictor or text conditioning.

Similar to SpeechFlow, we first pretrain the diffusion model on unlabeled audio data \cite{flowmatching}. We hypothesize this pretraining teaches the synthesizer how to speak in diverse voices, enabling it to learn the voice cloning task from minimal parallel S2ST data. For pretraining, we provide only masked context audio as conditioning. This makes our training task very similar to that of masked language modeling (MLM). We opt to have the diffusion synthesizer predict continuous EnCodec latent vectors \cite{encodec}, and then we use the EnCodec residual vector quantizer and decoder modules as a vocoder. At train time, we do not use the vocoder and compute an $L_2$ loss directly on the predicted EnCodec latents.

One of our main contributions is the mechanism by which we connect the diffusion synthesizer to the rest of the S2ST architecture without losing the benefits of pretraining, but still enabling zero-shot speaker preservation. We found a simple strategy to work well: we pass the upsampler's output through a small multilayer perceptron network, apply a positional encoding derived from the phoneme durations, and then add it to the context audio as conditioning. For the non-streaming S2ST task, context audio is just zeroes, as we are predicting an entire utterance from scratch. We note that the MLM-style training of the diffusion model facilitates further work on streaming S2ST, where we may want to continue speaking an already partially-spoken audio \cite{seamlessex}. We train the synthesizer to make use of its new conditioning by training the entire DiffuseST network on parallel S2ST data but freezing all non-synthesizer parameters.

\section{Experiments}
We evaluate DiffuseST on the task of zero-shot voice cloning S2ST from multiple languages into English. We evaluate the quality of the translation, the quality of the output audio, the similarity between the input and output speakers, and the model latency with both the NAT and diffusion synthesizers.

\subsection{Data}
To improve reproducibility, we are one of the first large-data-scale works in S2ST to train entirely on public data. We pretrain on several public audio corpora to circumvent the lack of parallel S2ST data. We list all the datasets we use below:
 \begin{itemize}
    \item \textbf{The People's Speech}: We use the clean and dirty train segments of The People's Speech dataset \cite{peoplespeech}, containing 52,000 hours of primarily English speech.
     \item \textbf{SpeechMatrix}: We take the en-en, es-en, fr-en, and de-en splits of the SpeechMatrix dataset \cite{speechmatrix}, containing 14569 hours of source audio. We transcribe all English audios with Whisper Medium \cite{whisper}.
     \item \textbf{LibriSpeech Multilingual}: We use an MT model to translate the es, fr, and de splits of the LibriSpeech Multilingual dataset (3959 hours of audio) into English \cite{libri}.
     \item \textbf{CVSS-T}: We use the high resource languages from the CVSS-T dataset (es-en, fr-en, de-en, and ca-en splits) \cite{cvss}, comprising 485 hours of source audio. CVSS-T has artificially synthesized speech labels that use a voice cloning TTS algorithm to preserve the voice of the input speaker in the output. We use the Montreal Forced Aligner to obtain duration labels for the target (English) phonemes \cite{mfa}.
 \end{itemize}

\subsection{Training Protocol}
\begin{table}[t]
\eightpt
  \caption{Parameter counts of network components.}
  \label{tab:params}
  \centering
  \begin{tabular}{l|c}
    \toprule
    \multicolumn{1}{l}{\textbf{Module}} & 
    \multicolumn{1}{c}{\makecell{\textbf{Parameter Count}}}  \\
    \midrule
    Acoustic Encoder & 88.2M \\
    Phoneme Decoder & 14.6M \\
    Attention & 1.0M \\
    \makecell[l]{NAT Synthesizer \\  \hspace{1.0cm} +Duration and Upsampling} & 44.6M \\
    \makecell[l]{Diffusion Synthesizer \\ \hspace{1.0cm} +Duration and Upsampling} & 102.1M \\
    \bottomrule
  \end{tabular}
\vspace{-15pt}
\end{table}

 We train our model in 4 stages. We focus on translating from Spanish, French, and German into English due to those languages' data availability. Our parameter counts are shown in Table \ref{tab:params}. We use an AdamW optimizer with a reduce-on-plateau learning rate schedule \cite{adamw}. When training on the smaller LibriSpeech and CVSS-T datasets, we apply SpecAugment and dropout to prevent overfitting \cite{specaugment}. 
 
 Our training curriculum proceeds as follows: 1) Pretrain the diffusion synthesizer on unlabeled audio from The People's Speech. This takes about 7 days using 8 A100 GPUs. 2) Pretrain the encoder and decoder on a speech-to-text translation (S2TT) task with SpeechMatrix using loss $\mathcal{L} = \mathcal{L}_p$. This also takes 7 days on 8 A100 GPUs. 3) Train the entire model with the NAT synthesizer on both S2TT and S2ST tasks simultaneously on a mixture of the LibriSpeech and CVSS-T datasets. We find that this multi-task training helps prevent overfitting on the smaller CVSS dataset and enhances translation quality. This step uses all losses: $\mathcal{L} = \mathcal{L}_p + \mathcal{L}_d + \mathcal{L}_s$ and takes 2-3 days on 8 A100 GPUs. 4) Replace the NAT synthesizer with the diffusion synthesizer and freeze all parameters other than the synthesizer. Finetune the model on just the CVSS-T dataset with loss $\mathcal{L} = \mathcal{L}_d + \mathcal{L}_s$. This step takes about 3 days on 8 A100 GPUs.

 We evaluate both the NAT synthesizer model produced by step 3 and the diffusion synthesizer model produced by step 4. Note that the encoder, decoder, attention, and duration modules of both models are the same because they are frozen in step 4.

\subsection{Translation Quality Results}

\begin{table}[h]
\eightpt
  \caption{Translation metrics computed on ASR transcriptions of model output on the high resource languages of CVSS-T. For BLEU and ChrF, higher is better. For WER and TER, lower is better. According to \cite{cvss}, references were generated with a PnGNAT voice-cloning TTS system using ground truth text translations.}
  
  
  \label{tab:trans}
  \centering
  \begin{tabular}{ c|l|cccc}
    \toprule
    \multicolumn{1}{c}{\makecell{\textbf{Source} \\ \textbf{Language}}} & 
    \multicolumn{1}{c}{\textbf{Method}} & 
                                         \multicolumn{4}{c}{\textbf{Translation Metrics}} \\
    \midrule
 & & BLEU & WER & ChrF & TER \\ \hline
 \multirow{2}{3em}{French} & NAT & 23.7 & 0.61 & 49.3 & 59.7 \\
 & Diffusion & 23.4 & 0.61 & 48.8 & 60.0 \\ \hline
  \multirow{2}{3em}{German} & NAT & 19.7 & 0.69 & 44.3 & 66.0 \\
  & Diffusion & 19.4 & 0.69 & 44.0 & 66.4 \\ \hline
  \multirow{2}{3em}{Spanish} & NAT & 23.9 & 0.63 & 50.9 & 60.7 \\
  & Diffusion & 23.1 & 0.65 & 50.3 & 62.8 \\
  \hline
  \multirow{2}{3em}{Catalan} & NAT & 20.3 & 0.69 & 47.0 & 66.9 \\
  & Diffusion & 20.1 & 0.69 & 46.9 & 67.1 \\ \hline
  \multirow{3}{3em}{Overall} & NAT & 21.9 & 0.65 & 47.9 & 63.3 \\
  & Diffusion & 21.7 & 0.66 & 47.5 & 64.0 \\
 
 & Reference & 90.8 & 0.06 & 95.5 & 5.5 \\ 
    \bottomrule
  \end{tabular}
  \vspace{-10pt}
\end{table}

 We evaluate DiffuseST on the test split of CVSS-T. Note that all speakers in the test set are unseen in the train set \cite{cvss}. In the decoder, we use a beam search with a size of $5$. For the diffusion synthesizer model, we use $25$ diffusion steps and a classifier-free guidance scale of $1.0$ at inference time. For the NAT synthesizer model, we use a HiFiGAN vocoder on the output to convert mel spectrograms to waveforms \cite{hifi}. To assess translation quality, we transcribe DiffuseST's output with Whisper-Large-V2 and compute several common translation evaluation metrics on the transcript by comparing to the text label in CVSS-T \cite{whisper}. Results are shown in Table \ref{tab:trans}. Since the encoder and decoder are the same for the NAT and diffusion synthesizer models, all differences in translation quality are attributable to pronunciation. 
 
 We observe that the diffusion synthesizer lags in this area by $0.25$ BLEU points across all languages, but this difference is not statistically significant ($p=.35$) and it potentially could be made up with further training on a larger dataset or hyper-parameter tuning. Regardless, we believe the dramatic gains in audio quality more than make up for the small tradeoff in pronunciation.
 
 As S2ST models in other works that are evaluated on CVSS use encoders in excess of 600M parameters, we lack direct comparisons in literature \cite{seamless, seamlessex, yejia2}. Unsurprisingly, we do not outperform these much larger works. For instance, SeamlessM4T-Large and AudioPalm achieve 36.5 and 32.5 BLEU respectively across all languages of CVSS-C. In future work, we plan on training versions of DiffuseST with higher parameter counts to create fair comparisons.

\subsection{Audio Quality and Speaker Similarity Results}
\begin{table}[b]
\eightpt
\vspace{-10pt}
  \caption{Mean audio quality metrics computed on model output using Torch Squim models. For both MOS and PESQ, higher is better. Dataset is high-resource languages of CVSS-T.}
  \label{tab:aud}
  \centering
  \begin{tabular}{ c|l|cc}
    \toprule
    \multicolumn{1}{c}{\makecell{\textbf{Source Language}}} & 
    \multicolumn{1}{l}{\textbf{Synth}} & 
                                         \multicolumn{2}{c}{\textbf{Audio Quality Metrics}} \\
    \midrule
 & & MOS & PESQ \\ \hline
 \multirow{2}{4em}{French} & NAT & 3.21 & 1.81 \\
 & Diffusion & \textbf{3.98} & \textbf{2.28} \\
  \hline
 \multirow{2}{4em}{German} & NAT & 3.20 & 1.83 \\
 & Diffusion & \textbf{3.98} & \textbf{2.29} \\
  \hline
 \multirow{2}{4em}{Spanish}  & NAT & 3.39 & 1.85 \\
 & Diffusion & \textbf{3.91} & \textbf{2.26} \\
 \hline
 \multirow{2}{4em}{Catalan} & NAT & 3.12 & 1.93 \\
 & Diffusion & \textbf{3.99} & \textbf{2.24} \\
  \hline
 \multirow{3}{4em}{Overall} & NAT & 3.23 & 1.85 \\
 & Diffusion & \textbf{3.97} & \textbf{2.27} \\
 & Reference & 4.02 & 3.11 \\
    \bottomrule
  \end{tabular}
\end{table}

Next, we evaluate the audio quality of DiffuseST with both the NAT and diffusion synthesizers using the Torch Squim models (results in Table \ref{tab:aud}) \cite{squim}. We find the diffusion synthesizer to outperform the NAT synthesizer in both MOS and PESQ scores by about 23\%. We believe that due to the diffusion model's extensive pretraining, it is better able to emulate the sounds of speech without producing excess artifacts or noise. Given the magnitude of the difference in MOS and PESQ scores, we think the diffusion synthesizer's gains in audio quality more than offset its issues with pronunciation, warranting further study of diffusion synthesizers for S2ST.

\begin{table}[t]
\eightpt
  \caption{Speaker similarity scores between input speech and generated speech as measured by cosine similarity between embeddings computed with the StyleTTS2 \cite{li2023styletts} style embedding network. Higher is better.}
  \label{tab:sim}
  \centering
  
  \begin{tabular}{ l|ccccc}
    \toprule
    \multicolumn{1}{l}{\textbf{Method}} & 
    \multicolumn{5}{c}{\makecell{\textbf{Source Language}}} \\
    \midrule
     & French & German & Spanish & Catalan & All \\ \hline
     NAT & .61 & .62 & .61 & .57 & .60 \\ 
     Diffusion & \textbf{.64} & \textbf{.64} & \textbf{.62} & \textbf{.63} & \textbf{.63} \\
     \hline
     Reference & .67 & .67 & .66 & .66 & .67 \\
    \bottomrule
  \end{tabular}
  \vspace{-15pt}
\end{table}

We also evaluate DiffuseST's voice cloning ability. We measure the cosine similarity between speaker embeddings of the input speech and model output. We use the style encoder from StyleTTS2 \cite{li2023styletts} to compute embeddings because of StyleTTS2's impressive voice cloning abilities. We observe that the diffusion synthesizer outperforms the NAT synthesizer with a 4.6\% improvement in cosine similarity across all languages. Again, we believe that the diffusion synthesizer's pretraining on diverse voices allows it to learn the complex voice-cloning task from little data. We believe this could be further improved through monolingual voice cloning training, generating more parallel S2ST data with a higher quality TTS algorithm, and potentially even training with backtranslation like Translatotron3 \cite{translatotron3}. We plan on exploring these methods in future work.

\subsection{Performance Results}
Lastly, we evaluate the speed of our model at inference time with both the NAT and diffusion synthesizers. We measure the mean inference time using a subset of data from the CVSS-T test set, running on a single A100 GPU using fp16 precision. We use flash attention \cite{flash} in the whisper encoder and diffusion synthesizer. The input audios used for performance evaluation have a mean duration of 5.61 seconds. We find the mean inference time of the model with the diffusion synthesizer to be 0.99 seconds while the NAT model takes 1.04 seconds, making both models over $5\times$ faster than real-time. Furthermore, despite having more than 2$\times$ more parameters, the diffusion synthesizer is faster than NAT one. In both cases, the most time-intensive step of inference is the decoder's beam search because it requires doing many forward passes, each of which must attend to the entire high-dimensional encoder output. We believe this low latency sets us up to adapt DiffuseST to work in a streaming setting.

 \section{Conclusion}
 We propose DiffuseST, a multilingual S2ST model featuring a non-autoregressive diffusion synthesizer capable of zero-shot voice cloning. In our experiments, we show our diffusion synthesizer improves audio quality and speaker preservation with minimal costs to pronunciation. We are one of the first large-data-scale S2ST works to train exclusively on public datasets, paving the way for more democratized S2ST work. DiffuseST is also one of the first direct, zero-shot speaker-preserving S2ST systems with fewer than 600M parameters, enabling over 5$\times$ faster than real-time processing. In the future, we intend to make DiffuseST work in a streaming setting by further optimizing the model latency while adding predictive capabilities to the phoneme decoder. We plan to continue to improve translation and audio quality via artificial dataset generation and backtranslation.

\ifinterspeechfinal
\fi

\balance
\bibliographystyle{IEEEtran}
\bibliography{mybib}

\end{document}